\documentclass[conference,a4paper]{IEEEtran}
\IEEEoverridecommandlockouts

\usepackage[hidelinks]{hyperref}
\usepackage[cmex10]{amsmath}%American Math Society(AMS) math formatting
\usepackage{amssymb,amsfonts}%AMS extra symbols and fonts
\usepackage{dblfloatfix}%fix double column figure ordering and placement

\usepackage[ruled,vlined]{algorithm2e}
\usepackage{graphicx}
\graphicspath{{Figures/PDF/}{Figures/PNG/}}

\usepackage{booktabs}
\usepackage{siunitx}
\usepackage[numbers,compress]{natbib}
\usepackage{texnames}
\usepackage{bm,bbm}
\usepackage{orcidlink}
\usepackage{listings}
\usepackage{tcolorbox}
\usepackage{tabularx}
\newcolumntype{C}{>{\centering\arraybackslash}X}

\makeatletter
\newcommand{\linebreakand}{%
  \end{@IEEEauthorhalign}
  \hfill\mbox{}\par
  \mbox{}\hfill\begin{@IEEEauthorhalign}
}
\makeatother

\begin{document}

\title{\uppercase{SSL4EO-S12 v1.1: A Multimodal, Multiseasonal Dataset for Pretraining, Updated}
\thanks{Embed2Scale is co-funded by the EU Horizon Europe program under Grant Agreement No.\ \text{101131841}. Additional funding for this project has been provided by the Swiss State Secretariat for Education, Research and Innovation (SERI) and UK Research and Innovation (UKRI).  We thank Yi Wang for valuable discussions, a key architect of the SSL4EO-S12 dataset.}
}

\author{
    \IEEEauthorblockN{Benedikt Blumenstiel}
	\IEEEauthorblockA{\textit{IBM Research Europe}\\
		benedikt.blumenstiel@ibm.com}
	
    \and
    
	\IEEEauthorblockN{Nassim Ait Ali Braham}
	\IEEEauthorblockA{\textit{German Aerospace Center}\\
		nassim.aitalibraham@dlr.de}
	
    \and
	
    \IEEEauthorblockN{Conrad M Albrecht}
	\IEEEauthorblockA{\textit{German Aerospace Center}\\
		conrad.albrecht@dlr.de}
        
    \linebreakand
    
	\IEEEauthorblockN{Stefano Maurogiovanni}
	\IEEEauthorblockA{\textit{Julich Supercomputing Centre}\\
        \textit{University of Iceland}\\
		s.maurogiovanni@fz-juelich.de}
    
    \and
	
    \IEEEauthorblockN{Paolo Fraccaro}
	\IEEEauthorblockA{\textit{IBM Research Europe}\\
		paolo.fraccaro@ibm.com}
}

\maketitle

\begin{abstract}
   This work presents SSL4EO-S12~v1.1, a multimodal, multitemporal Earth Observation dataset designed for pretraining large-scale foundation models. Building on the success of SSL4EO-S12, this extension updates the previous version to fix geospatial alignment inaccuracies and the inefficent data structure. The dataset allows low-barrier, analysis-ready data loading while maintaining the predecessor's spatial coverage of the world’s 10,000 largest cities and surrounding geographies, resulting in 246k time series with nearly one million image patches. We package each time series in Zarr file format stored in WebDataset tar shards for efficient data loading and representation of meta-information such as cloud masks. We add  new modalities for elevation, land-cover, and vegetation to support multimodal pre-training. Released under the CC-BY-4.0 license, SSL4EO-S12~v1.1 facilitates open research and provides a robust foundation for future advancements in self-supervised learning and geospatial analysis. The dataset is available online through \url{https://huggingface.co/datasets/embed2scale/SSL4EO-S12-v1.1}.
\end{abstract}

\section{Introduction}
\label{sec:intro}

\begin{figure*}[bht!]
    \centering
    \includegraphics[width=\linewidth]{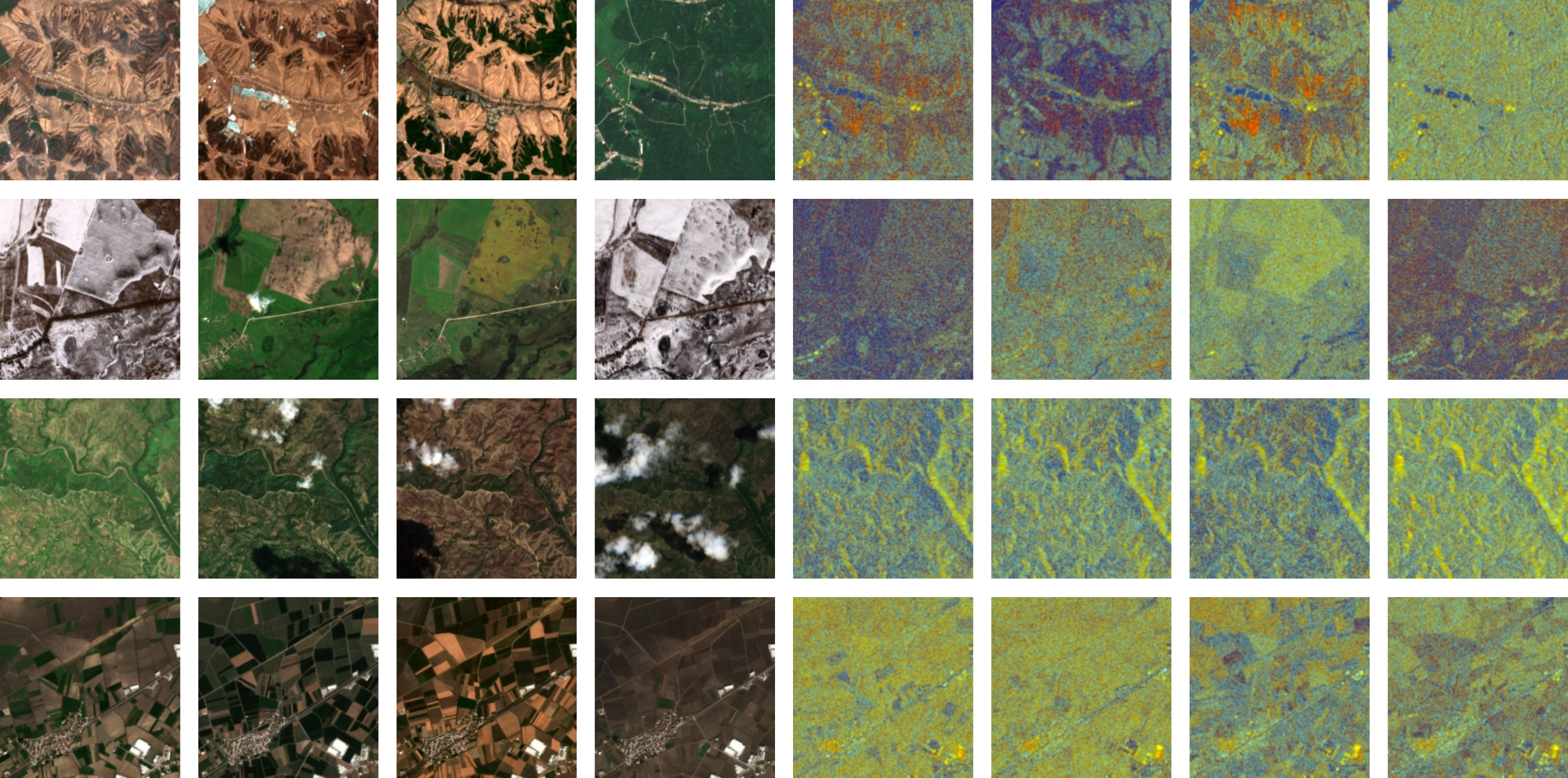}    
    \caption{Example patches from SSL4EO-S12~v1.1 illustrating four columns with seasonal timestamps of Sentinel-2 (S-2) L2A (left) and Sentinel-1 (S-1) GRD (right) products. S-1 GRD is visualized using VV-VH-VV/VH pseudo coloring.}
    \label{fig:SSL4EO-S12_samples}
\end{figure*}

Large-scale Earth Observation~(EO) datasets play a pivotal role in the development of Foundation Models~(FM)~\cite{ssl4eos12,terramind,decur,prithvi2,dofa,croma} for remote sensing. By training on diverse and extensive collections of remote sensing imagery, these models achieve better accuracy and converge faster~\cite{prithvi2,dofa}. Among such datasets, SSL4EO-S12~\cite{ssl4eos12} has been widely adopted and enabled the development of EO FMs such as CROMA and DeCur~\cite{decur,croma}. The dataset includes around 250k global locations with four seasonal images covering urban and rural landscapes.

Despite its success, the original SSL4EO-S12 dataset presented two main challenges: (1)~spatial misalignments between Sentinel-1~(S-1) and Sentinel-2~(S-2) patches and (2)~limited availability of analysis-ready data~(ARD). To address these issues, we present SSL4EO-S12~v1.1. We ensure proper alignment of the modalities by downloading larger areas and reprojecting S-1 GRD data to match S-2 L2A coordinates. Furthermore, we introduce stricter filtering of \texttt{NaN} values and provide the dataset in an ARD format suitable for direct pretraining of EO FMs. Figure~\ref{fig:SSL4EO-S12_samples} illustrates S-2 L2A and S-1 GRD time series samples from SSL4EO-S12~v1.1. Furthermore, we release additional modalities to support multimodal pre-training, including land-use land-cover~(LULC) maps, elevation data~(DEM) and a vegetation index~(NDVI) aligned with the TerraMesh dataset~\cite{terramesh}.

Our contributions include (1)~updating the original SSL4EO-S12 locations with aligned satellite patches, (2)~strict filtering and preprocessing in an ARD-format, and (3)~releasing additional aligned modalities. 
%This paper is organized as follows: We first detail the data processing steps necessary to construct SSL4EO-S12~v1.1, including alignment, reprojection, and filtering. We then present an overview of the final dataset and discuss its applicability for EO FM training pipelines.

\section{Related Work}
\label{sec:related_work}

The development of large-scale, multimodal Earth Observation datasets has been critical for advancing self-supervised learning in remote sensing. Existing datasets such as BigEarthNet-MM~\cite{bigearthnetmm} and SEN12MS~\cite{sen12ms} have provided valuable baselines, yet they often focus on single timestamps, limiting their utility for pre-training foundation models that require diverse spatiotemporal inputs~\cite{prithvi2}.

SSL4EO-S12 and its extensions~\cite{wang2025unifiedcopernicusfoundationmodel,ssl4eo_eu_forest} addressed many of these limitations by including multiple modalities and covering multiple seasons, thereby supporting a broader range of downstream tasks~\cite{ssl4eos12}. Several state-of-the-art models have leveraged SSL4EO-S12 to achieve improvements in accuracy and generalizability~\cite{ssl4eos12,terramind,decur,croma}. For instance, contrastive learning approaches and transformer-based architectures have demonstrated notable performance gains when pre-trained on SSL4EO-S12.
Recently, datasets such as TerraMesh~\cite{terramesh} and MMEarth~\cite{mmearth} have introduced additional modalities for multimodal pre-training. However, they lack time series data, which is required to train multitemporal models.

\section{Data Processing}
\label{sec:processing}

\begin{figure*}[tbh]
    \centering
    \includegraphics[width=0.95\textwidth]{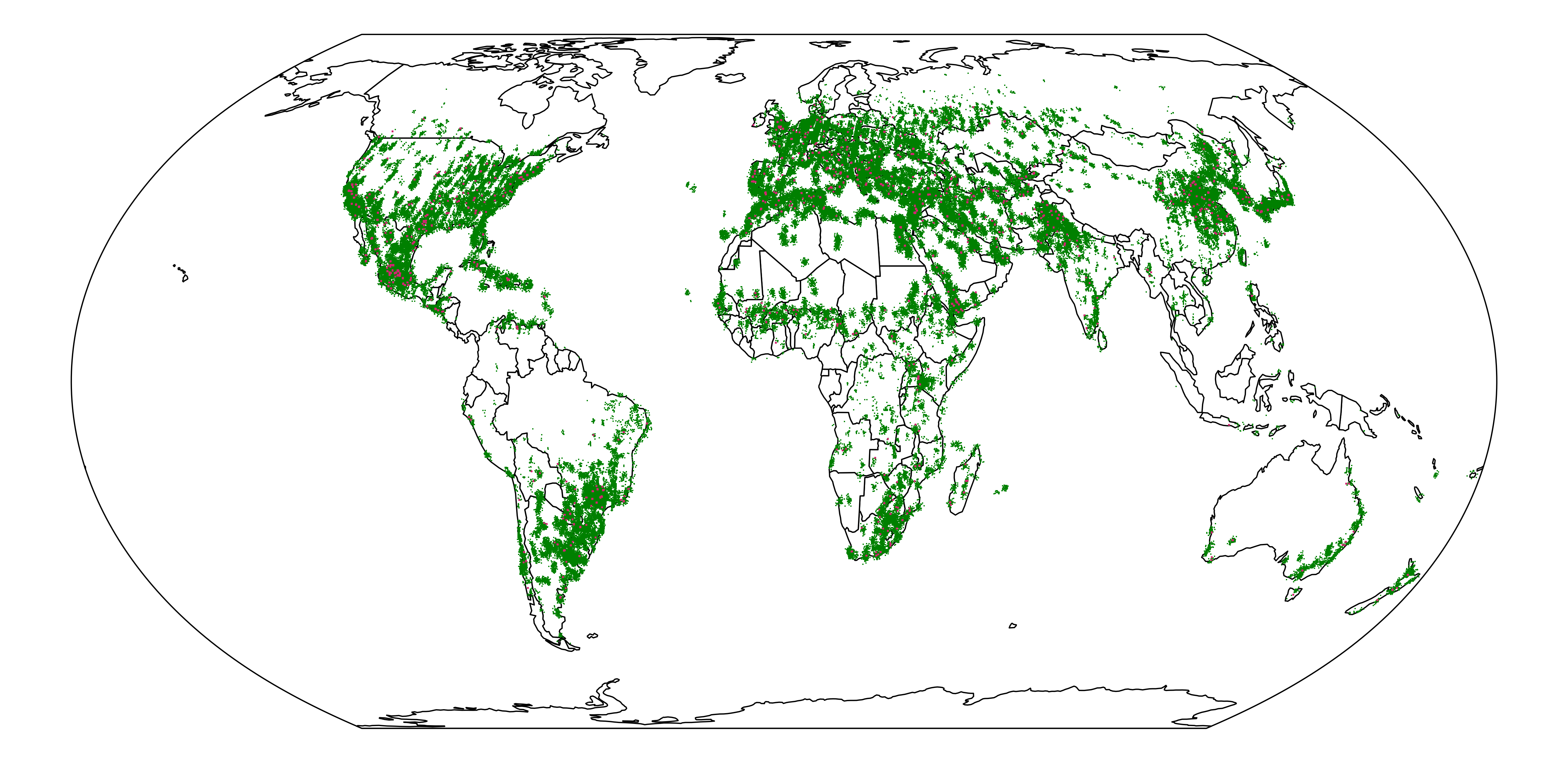}
    \caption{Global distribution of SSL4EO-S12~v1.1 training (green) and validation (magenta) samples (size not to scale).}
    \label{fig:map}
\end{figure*}

All data for SSL4EO-S12~v1.1 were acquired from Google Earth Engine~\cite{gee}. In SSL4EO-S12, all samples were projected onto EPSG:4326, which can distort distances, especially near the poles. In v1.1, we retain native UTM zones to preserve spatial accuracy and avoid unnecessary reprojections. For each location in v1, four timestamps from different seasons are selected from 2019 through 2021 with cloud coverage below 10\%~\cite{ssl4eos12}. We select the timestamp closest to the S-2 timestamps in v1.
To ensure correct spatial alignment, we first download a larger area of interest for both S-2 and S-1 to buffer spatial mismatch on reprojection of these modalities. Next, we verify that all S-2 timestamps for each location share the same UTM zone. If timestamps spanned multiple zones—common at UTM boundaries, we pick the majority CRS for that location, crop a reference timestamp to a common size of 264x264 pixels, and reproject the remaining timestamps accordingly. Subsequently, we project S-1 patches onto the curated 264x264-pixel S-2 patches. 
Following v1~\cite{ssl4eos12}, we fix the patch sizes to 264 pixels because it is divisible by six, accommodating S-2 pixel resolutions of 10\,m, 20\,m, and 60\,m. All bands are upsampled to 10\,m using nearest interpolation. 

Next, we drop any location with more than 1\% of missing values (\texttt{NaN}) for any timestamp or channel. The remaining \texttt{NaN}s are filled in using nearest neighbor interpolation. We split the dataset into training and validation sets, following a 99\%-to-1\% partition defined by TerraMesh~\cite{terramesh} to avoid spatial overlap that may leak information while training models.

The cloud masks provided by S-2 products are prone to errors. Therefore, they are used for data sampling only. Following~\cite{majortom}, we use the SEnSeI v2 model~\cite{senseiv2} to provide accurate cloud masks for SSL4EO-S12~v1.1. 

Furthermore, we provide S-2 RGB data based on L2A products. 
For data compression reasons, often reflectance value ranges of 0--2000 are mapped to 0--255 (single-byte integers). However, this approach may lead to saturation in bright regions (e.g., clouds, snow, and sand). A 0--max mapping can lead to very dark images. To address this issue, we first determine the 2\% and 98\% quantiles over the RGB channels within each image and multiply a factor of 0.5 to values outside these bounds to reduce extreme pixel intensities.
We then map these adjusted values to an 8-bit (0--255) range, using the 0.2\% and 99.8\% quantiles as the lower and upper limits, respectively. If the computed upper limit falls below 2000, we clamp it to 2000. Finally, if the median of the RGB values in an image is below 1000, we set the lower limit to 0 to avoid further darkening already dim images. 
% The exact transformation is provided in Listing~\ref{lst:rgb}.
%
% \begin{figure}
% \centering
% \vspace{-3mm}
% \begin{lstlisting}[label={lst:rgb}, caption=Pseudocode for the S-2-to-RGB transformation.]
% \end{lstlisting}
% \begin{tcolorbox}[boxrule=0pt, sharp corners, width=0.98\linewidth, boxsep=3pt, left=5pt, right=3pt, top=0pt, bottom=0pt]
% \begin{lstlisting}[language=Python]
% import numpy as np
%
% def s2_to_rgb(img):
%     img = img[[3,2,1]]  # select RGB
%     Q2, Q98 = np.quantile(img, [0.02, 0.98])
%     img = np.where(img >= Q2, img, 
%                    Q2 + (img - Q2) * 0.5)
%     img = np.where(img <= Q98, img, 
%                    Q98 + (img - Q98) * 0.5)    
%     Q02, Q50, Q998 = np.quantile(img, 
%         [0.002, 0.5, 0.998])
%     U = max(2000, Q998)
%     L = 0 if Q50 < 1000 else Q02
%     img = (img - L) / (U - L) * 255
%    
%     img = np.clip(img, 0, 255)
%     return img.astype(np.uint8)
%
% \end{lstlisting}
% \end{tcolorbox}
% \vspace{-3mm}
% \end{figure}
%
This pipeline ensures a more visually balanced and artifact-free RGB representation of the S-2 data, as visible in Figure~\ref{fig:SSL4EO-S12_samples}.

The additional modalities are sourced from TerraMesh~\cite{terramesh} and reshaped to match the four-timestamp time series. NDVI was computed from Sentinel-2 L2A reflectances using the standard $(B08 - B04) / (B08 + B04)$ formula and serves as a proxy for vegetation tasks. Land-use/land-cover maps were sourced from ESRI
%\footnote{ESRI LULC source: \url{https://planetarycomputer.microsoft.com/dataset/io-lulc-annual-v02}}
~\cite{esrilulc} and augmented with SEnSeI v2 cloud and snow masks to improve temporal consistency; as LULC is yearly, it remains nearly static across the time series. The Copernicus DEM
%\footnote{Copernicus DEM source: \url{https://dataspace.copernicus.eu/explore-data/data-collections/copernicus-contributing-missions/collections-description/COP-DEM}}
~\cite{copernicusdem} provides elevation context at 30\,m resolution and was bilinearly resampled to the Sentinel-2 grid. Unlike the other modalities, DEM is included as a single timestamp since terrain does not change over time. All modalities were co-registered to the Sentinel-2 grid at 10\,m resolution and stored in Zarr format.

For efficient data access and storage, we stack the samples into Zarr files (\texttt{ZipStore}, Zarr version~2\footnote{Zarr v2: \url{https://zarr-specs.readthedocs.io/en/latest/v2/v2.0.html}}) with four timestamps each. The samples are shuffled and stored in tar shards to reduce the total file count, an important consideration for storing data on an HPC cluster or HuggingFace. By using tar files, 
the dataset can directly be streamed with \texttt{WebDataset}\footnote{WebDataset documentation: \url{https://github.com/webdataset/webdataset}}. 
Similar to TerraMesh~\cite{terramesh}, we save modalities in seperate folders instead of shared tar files which is natively expected by WebDataset. This makes working with single modalities feasible and we provide code to adapt the WebDataest data loading.
Alternatively, the data can be unpackaged and used with \texttt{xarray} and it supports lazy data loading via \texttt{Dask}. The data is chunked by timestamp to enable efficient loading of individual time steps if no time series is needed. The S-2 bands are stored as 16-bit (2 bytes) integers, and the S-1 data is stored as 16-bit floats. Combined with Zarr's built-in compression, these measures reduce storage overhead.

\begin{figure*}[tbh]
    \centering
    \includegraphics[width=\textwidth]{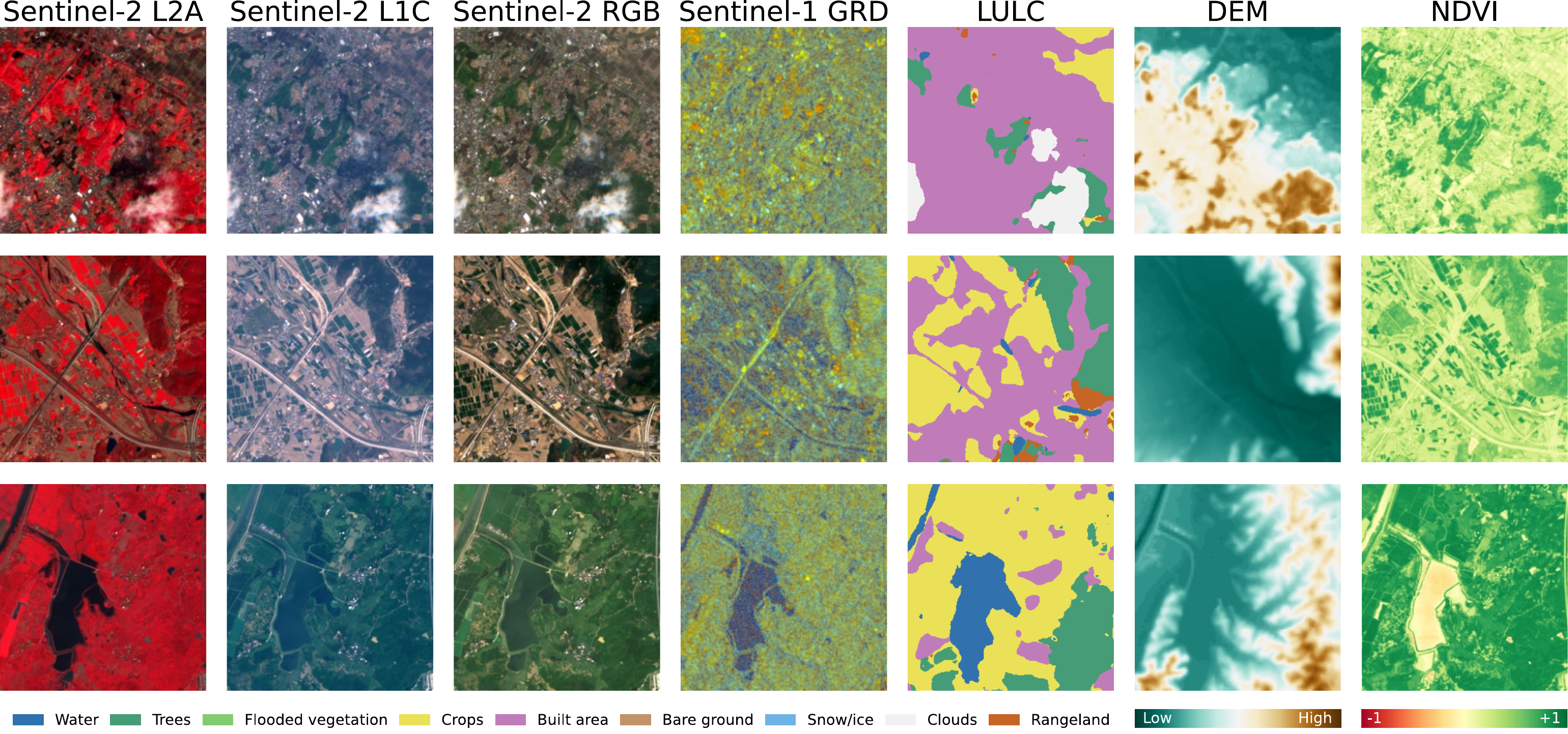}
    \caption{Examples of all spatio-temporally aligned modalities in SSL4EO-S12~v1.1 using a single timestep. Sentinel-2 L2A uses IRRG pseudo-coloring, and Sentinel-1 GRD is visualized in dB scale as VV-VH-VV/VH. Copernicus DEM is scaled based on the image value range with an additional 10\,meter buffer. 
    %Each sample is a time series of four seasons per modality which are not visualized.
    }
    \label{fig:modalities}
\end{figure*}

\section{Dataset}
\label{sec:data}

The final SSL4EO-S12~v1.1 dataset (ARD version) contains 246,144 distinct locations, each with four timestamps, resulting in a total of 984,576 samples. Figure~\ref{fig:map} visualizes the global distribution of these samples. We release the dataset on Hugging Face at \url{https://huggingface.co/datasets/embed2scale/SSL4EO-S12-v1.1} under a permissive CC-BY-4.0 license.

\begin{table}[tbh]
    \centering
    \small
    \setlength{\tabcolsep}{3pt}
    \caption{Summary of modalities included in SSL4EO-S12~v1.1. The range includes the typical value range, not fixed bounds.}
    \label{tab:modalities}
    \begin{tabularx}{\linewidth}{lcCCC}
    \toprule
    \textbf{Modality} & \textbf{Type} & \textbf{Range} & \textbf{Units} & \textbf{\# Bands} \\
    \midrule
    Sentinel-1 GRD   & SAR & -50 -- +1 & dB & 2 \\
    Sentinel-2 L1C   & Optical   & 0 -- 10k & DN & 13 \\
    Sentinel-2 L2A   & Optical   & 0 -- 10k & DN & 12 \\
    Sentinel-2 RGB   & Optical   & 0 -- 255  & -- & 3 \\
    NDVI & Index & -1 – +1 & -- & 1 \\
    Copernicus DEM & Elevation & 0 – 8k & Meter & 1 \\
    LULC & Annotation & 0 – 9 & Classes & 1 \\
    \bottomrule
    \end{tabularx}
\end{table}

As in v1, the dataset covers all four seasons, including snowy imagery. Sampling within a 50\,km radius of the world’s 10,000 most populated cities ensures near-global coverage while focusing on urban areas but also includes ocean patches and rural areas. This diversity makes EO FMs pre-trained on SSL4EO-S12~v1.1 suitable for numerous downstream tasks, such as land use and change detection, segmentation of floods, or object detection for urban planning. The dataset also supports the training of multi-temporal models as well as multimodal fusion techniques. Table~\ref{tab:modalities} includes further details about the modalities and Figure~\ref{fig:modalities} shows examples of single timestamps with all modalities. Providing S-2 L1C, S-2 L2A, and S-2 RGB ensures that pre-trained models can be used for many downstream tasks regardless of top-of-atmosphere (L1C), bottom-of-atmosphere (L2A) or RGB-only datasets.

We provide an overview of typical Zarr Zip files and data loading code in the \href{https://github.com/DLR-MF-DAS/SSL4EO-S12-v1.1}{git repository}.
Each file contains a sample with four timestamps, sorted by date rather than season to vary the inputs. Relevant metadata are stored with the image data as additional variables. The metadata includes center coordinates in latitude and longitude (EPSG:4326), pixel coordinates in the corresponding UTM zones, the file identifiers, as well as timestamps. 
The SEnSeI v2 cloud masks include the classes: \textit{land}, \textit{water}, \textit{snow}, \textit{thin cloud}, \textit{thick cloud}, \textit{cloud shadow}, and \textit{no data}. The shared storage in one Zarr file simplifies data management and allows users to efficiently reference image location and timestamps.

\section{Experimental Results}

The newly added modalities—LULC, DEM, and NDVI—introduce semantic and topographic context beyond raw spectral and radar data. These layers can serve as proxy tasks for self-supervised pre-training, as demonstrated in TerraMesh~\cite{terramesh} and MMEarth~\cite{mmearth}, where semantic annotations improve representation learning. Unlike these datasets, SSL4EO-S12~v1.1 enables training multitemporal models by providing spatio-temporally aligned time series. 

\begin{figure}[tbh]
    \centering
    \vspace{-3mm}
    \includegraphics[width=\linewidth]{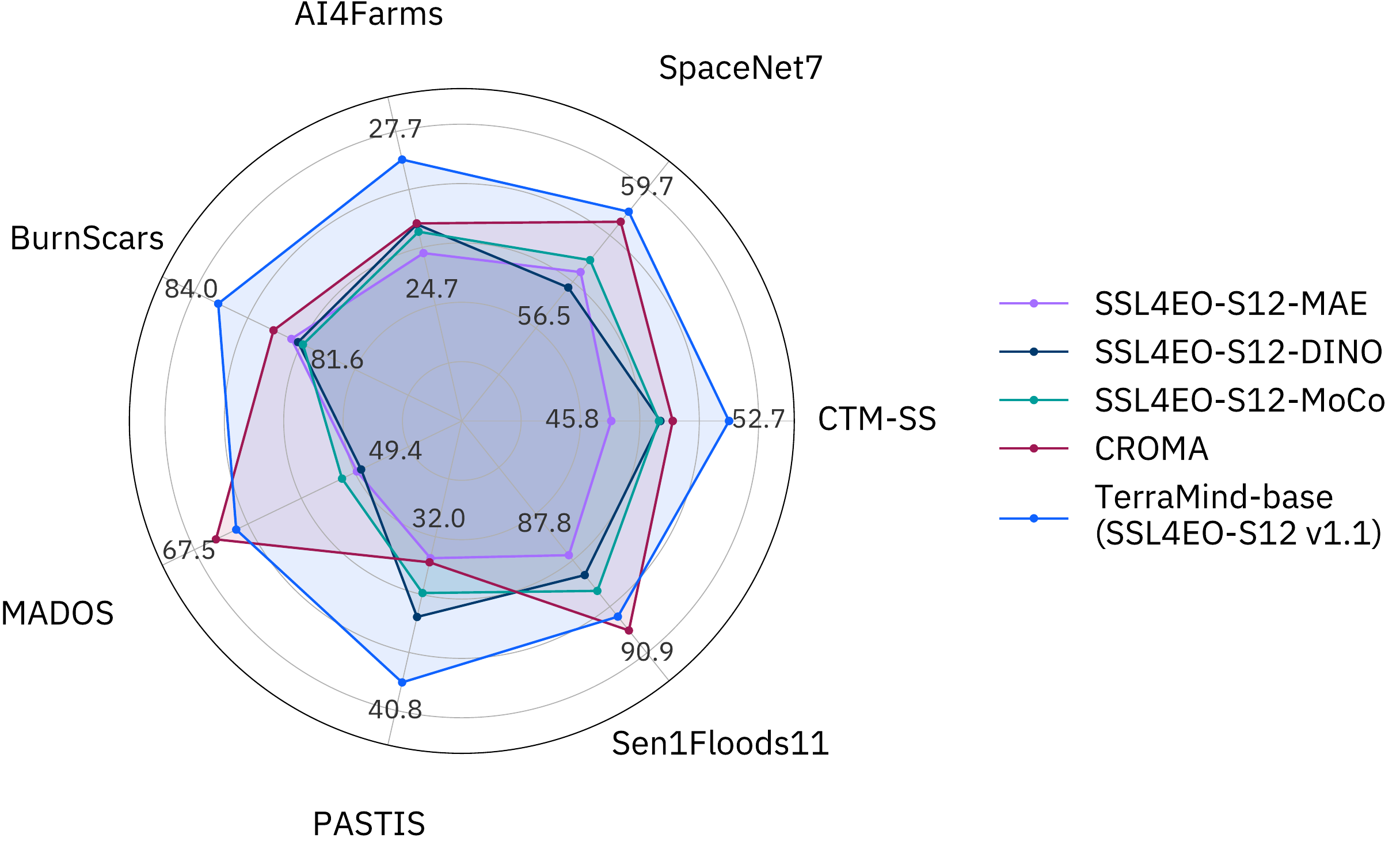}
    \caption{Comparison on PANGAEA between v1 models trained only on S-2 (MAE, DINO, and MoCo) with CROMA, trained on v1 S-1 and S-2, as well as TerraMind, trained with multimodal v1.1 data. Results taken from~\cite{pangaea,terramind}.}
    \label{fig:radar}
\end{figure}

Figure~\ref{fig:radar} shows a comparison on the PANGAEA benchmark~\cite{pangaea} across multiple models, with results sourced from~\cite{terramind,pangaea}. We display only the subset of benchmark tasks for which all sources report results. CROMA~\cite{croma} is trained on S-1 and S-2 data and already outperforms the FM adaptations of MAE, DINO, and MoCo trained only on S-2 data~\cite{ssl4eos12}. The multimodal TerraMind model~\cite{terramind} trained on SSL4EO-S12~v1.1 with multiple modalities performs best on all datasets except for two. The multimodal data in v1.1 directly enabled the token prediction task used by TerraMind.

\section{Conclusion}

SSL4EO-S12~v1.1 extends the original dataset into a multimodal, multitemporal resource tailored for pretraining of EO foundation models. By addressing spatial misalignment and introducing an analysis-ready format, it simplifies integration into modern training pipelines. The addition of semantic and topographic modalities enables richer representations through self-supervised learning. Combined with efficient Zarr packaging and WebDataset streaming, this dataset provides an easy-to-use resource for developing and experimenting with the next-generation of EO models. SSL4EO-S12~v1.1 joins the family of SSL4EO-$\ast$, open-source datasets to serve the EO data science community for research on multimodal fusion, temporal reasoning, and geospatial representation learning.

\small
\bibliographystyle{IEEEtranN}
\bibliography{references}

\end{document}